\documentclass[nohyperref]{article}

\usepackage{microtype}
\usepackage{graphicx}
\usepackage{subfigure}
\usepackage{booktabs} 

\usepackage{multicol} 
\usepackage{multirow}
\usepackage{algorithm}
\usepackage{setspace}
\usepackage{hyperref}
\usepackage{amsthm,amsmath,amssymb}
\usepackage{mathrsfs}

\usepackage[accepted]{icml2022}

\usepackage{amsmath}
\usepackage{amssymb}
\usepackage{mathtools}
\usepackage{amsthm}

\usepackage{color, colortbl}
\definecolor{Gray}{gray}{0.9}

\usepackage[capitalize,noabbrev]{cleveref}

\theoremstyle{plain}

\theoremstyle{definition}

\theoremstyle{remark}

\usepackage[textsize=tiny]{todonotes}

\icmltitlerunning{VLMixer: Unpaired Vision-Language Pre-training via Cross-Modal CutMix}

\begin{document}

\twocolumn[
\icmltitle{VLMixer: Unpaired Vision-Language Pre-training via Cross-Modal CutMix}




\begin{icmlauthorlist}
\icmlauthor{Teng Wang}{sustech,hku}
\icmlauthor{Wenhao Jiang}{tencent}
\icmlauthor{Zhichao Lu}{sustech}
\icmlauthor{Feng Zheng}{sustech}
\icmlauthor{Ran Cheng}{sustech}
\icmlauthor{Chengguo Yin}{tencent}
\icmlauthor{Ping Luo}{hku}
\end{icmlauthorlist}

\icmlaffiliation{sustech}{Department of Computer Science and Engineering, Southern University of Science and Technology}
\icmlaffiliation{tencent}{Data Platform, Tencent}
\icmlaffiliation{hku}{Department of Computer Science, The University of Hong Kong}

\icmlcorrespondingauthor{Feng Zheng}{f.zheng@ieee.org}

\icmlkeywords{Machine Learning, ICML}

\vskip 0.3in
]



\printAffiliationsAndNotice{}  

\begin{abstract}

Existing vision-language pre-training~(VLP) methods primarily rely on paired image-text datasets, which are either annotated by enormous human labors, or crawled from the internet followed by elaborate data cleaning techniques. 
To reduce the dependency on well-aligned image-text pairs, it is promising to directly leverage the large-scale text-only and image-only corpora. 
This paper proposes a data augmentation method, namely cross-modal CutMix (CMC), for {implicit} cross-modal alignment learning in unpaired VLP. 
Specifically, CMC transforms natural sentences from the textual view into a multi-modal view, where visually-grounded words in a sentence are randomly replaced by diverse image patches with similar semantics. 
There are several appealing proprieties of the proposed CMC. 
First, it enhances the data diversity while keeping the semantic meaning intact for tackling problems where the aligned data are scarce; 
Second, by attaching cross-modal noise on uni-modal data, it guides models to learn token-level interactions across modalities for better denoising. 
Furthermore, we present a new unpaired VLP method, dubbed as VLMixer, that integrates CMC with contrastive learning to pull together the uni-modal and multi-modal views for better instance-level alignments among different modalities. 
Extensive experiments on five downstream tasks show that VLMixer could surpass previous state-of-the-art unpaired VLP methods. 
{Project page: \url{https://github.com/ttengwang/VLMixer}}
\end{abstract}

\vspace{-2em}

\section{Introduction}

\begin{figure}[t]
    \centering
    \includegraphics[width=0.50\textwidth]{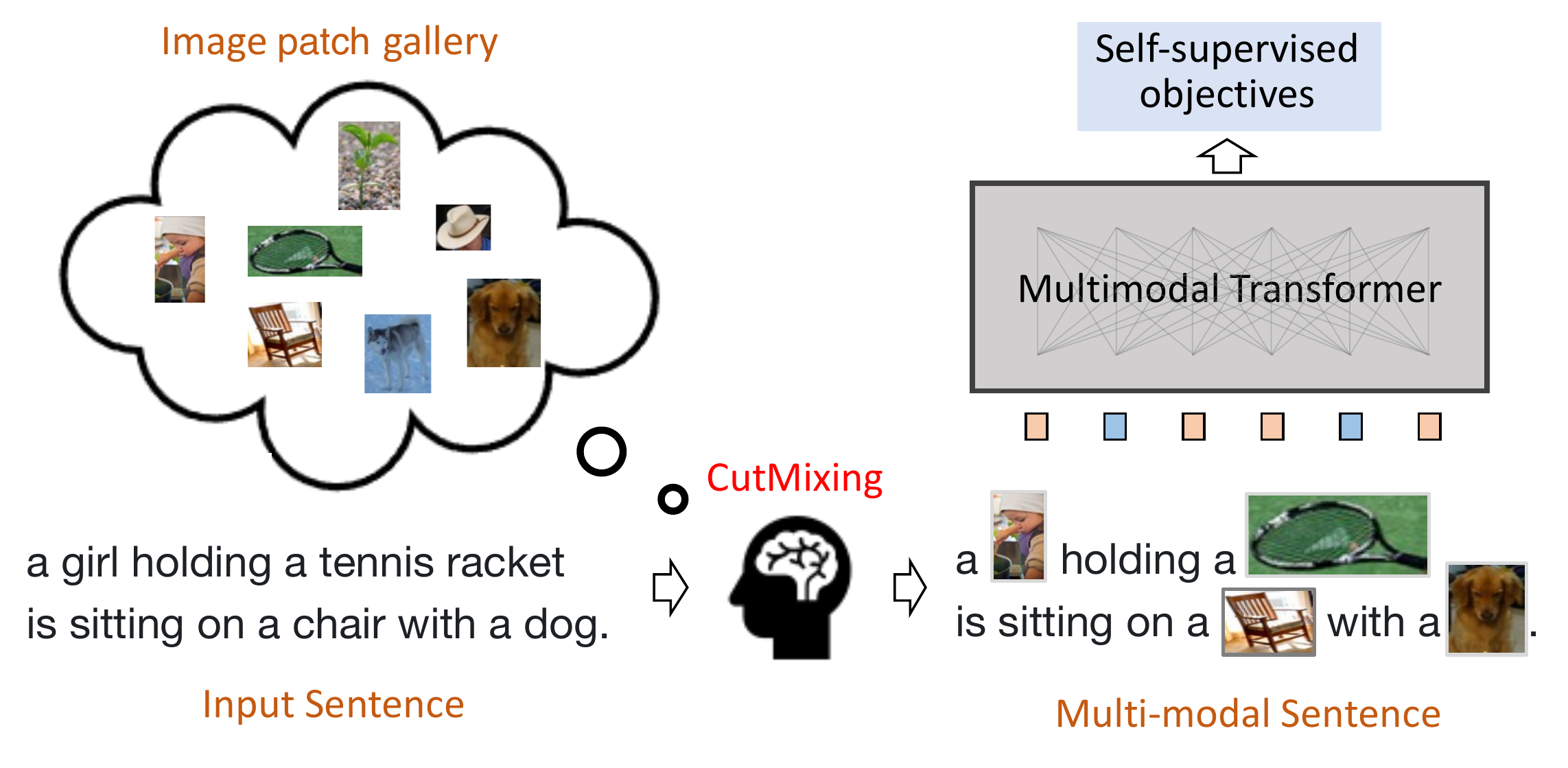}
    \caption{Illustration of the cross-modal CutMix (CMC). By randomly replacing the grounded words in a sentence with visual tokens, 
    we obtain diverse ``multi-modal sentences" without changing the semantics but injecting cross-modal noises.
    }
    \label{fig: fig1}
    \vspace{-1em}
\end{figure}

Vision-language pre-training (VLP) has received increasing attention and brought real benefits to a large variety of downstream tasks in the recent past~\cite{tan2019lxmert,li2019visualbert,lu2019vilbert,chen2019uniter,li2020oscar, cao2020behind,hu2020vivo,li2020unimo,zhang2021vinvl,radford2021learning,kim2021vilt,li2021align,jia2021scaling}. 
The success of existing VLP models mainly comes from manually-labeled and well-aligned image captioning datasets, such as COCO~\cite{lin2014microsoft} and Visual Genome~\cite{krishna2017visual}, and high-capacity transformer models~\cite{vaswani2017attention} with effective pre-training objectives for discovering the cross-modal alignments.
In mainstream VLP methods, the modeling of cross-modal alignment has been proved to be effective in achieving promising performance for several downstream tasks~\cite{cao2020behind}. 
At a global level, image-text matching losses~\cite{li2020oscar,chen2019uniter,zhang2021vinvl} are designed to guide the model to judge whether the input image and sentence are aligned. 
With the warranty of the instance-level alignment, the self-attention layers could further excavate the fine-grained interactions between input tokens across two modalities in an implicit manner. 

Although promising performance has been reported, the improvements of these methods that pre-trained on well-aligned datasets have gradually reached saturation due to the cost of annotating large-scale datasets. 
The following works alleviate this issue by introducing weakly-aligned image-caption pairs, which contain noisy annotations but are easy to access and scale-up. 
Unpaired vision-language pre-training~\cite{li2021unsupervised} further relieves the reliance on paired image-caption data, aiming to learn multi-modal representation from the standalone image and text corpus. 

Without explicit annotations of the cross-modal correspondence, unpaired VLP faces the challenge of distinguishing the alignment degree between an image and a text effectively. 
Previous work~\cite{li2021unsupervised} utilizes a shared encoder to learn a joint representation space, meanwhile introducing image tags as an intermediate representation to bridge the two modalities.
We argue that, image tags are not {\color{black} reliable}
representations for complex images, as the permutation-invariant nature and the lack of syntactic structure make them unrecognizable for visual relationships between objects. This further hurts downstream tasks that heavily rely on fine-grained alignments between images and texts, such as NLVR$^2$~\cite{suhr2018corpus} and image-text retrieval.

For fine-grained alignments across modalities, we propose the cross-modal CutMix (CMC) to construct a new representation, ``multi-modal sentence", to connect images and texts, which not only preserves the linguistic nature of a sentence but also links to the visual elements in images. 
A natural sentence can be transformed into its multi-modal view by replacing some grounded words with the image patches of the same semantic meaning\footnote{We assume that the text corpus shares a proportion of visual concepts with the image dataset, as it is unpractical to align arbitrary uni-modal datasets with semantic disparity, such as aligning cooking images with a corpus of mathematical terms.}. 
To this end, we create a visual patch gallery with diverse visual patterns from the image-only datasets, where high-quality visual patches are detected and tagged by a concept detector. 
As shown in Fig.~\ref{fig: fig1}, the input sentence after cutmixing not only preserves the syntactic and semantic information but also introduces the visual tokens as the cross-modality noise. 
Together with the mask-then-predict training objectives, it is promising for the model to learn cross-modal interactions among input tokens and token-level alignment between ``grounded words" and image patches.

Furthermore, we propose a contrastive learning framework to fully exploit the instance-level alignments between modalities. 
For an input sentence, CMC could produce a multi-modal view of the sentence, which has the same semantics as the language view. 
The contrastive supervision is then adopted to pull together the semantic-similar instances with different views and push away semantically different instances from the anchor. 
By distinguishing the positive samples from negative samples, the model could judge the alignment between inputs with different modalities. 
 
\begin{figure}[t]
    \centering
    \includegraphics[width=0.5\textwidth]{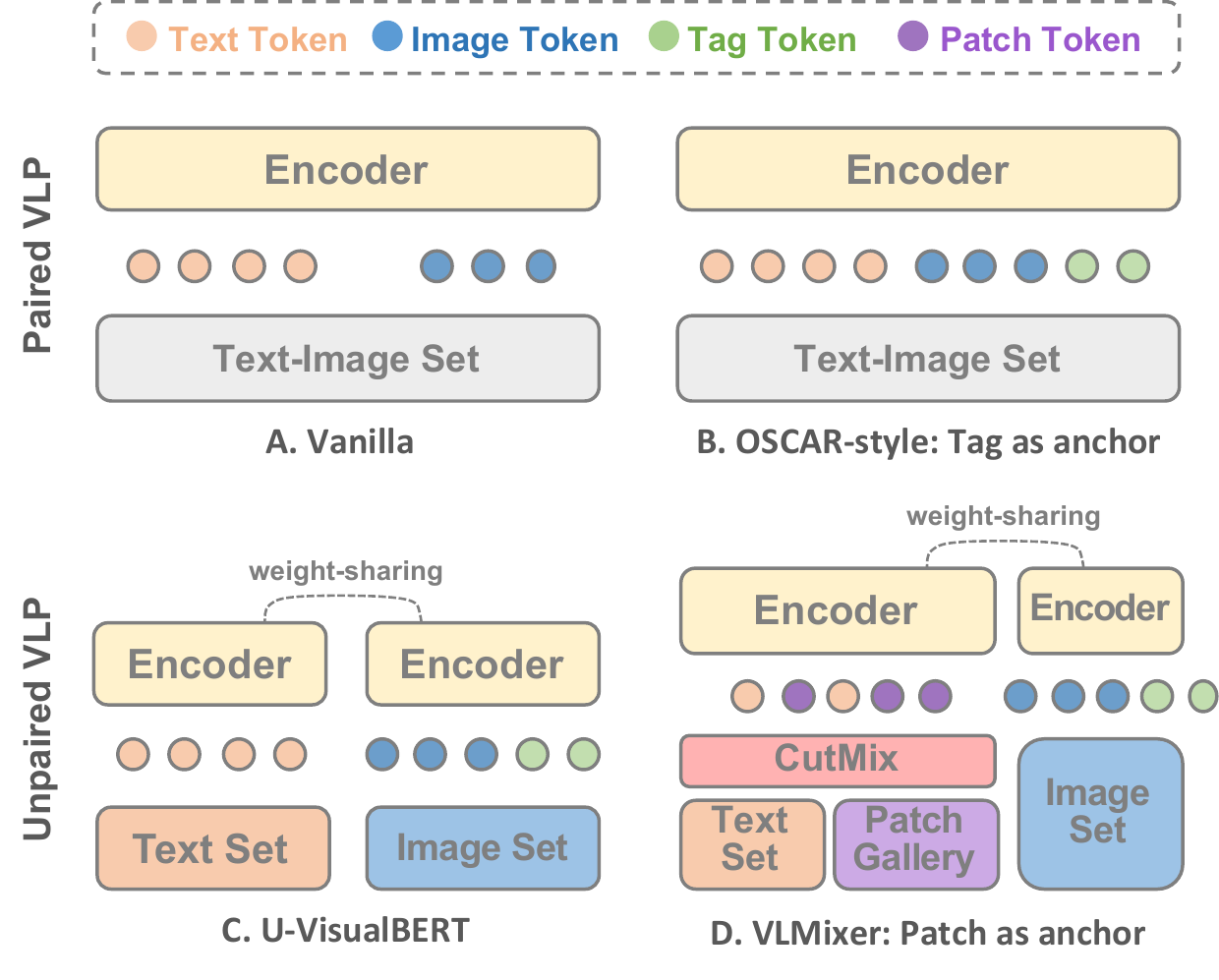}
     \caption{Comparison between existing methods and our framework in model structure and token construction. (A) Vanilla-style methods~\cite{chen2019uniter,tan2019lxmert,li2019visualbert} directly concatenate the visual tokens (object or grid features) with paired language tokens as inputs. (B) Oscar-style methods~\cite{li2020oscar, zhang2021vinvl} utilize the image tags extracted by an object detector, serving as the anchor points that existed in both visual and text data to bridge two modalities for better alignment learning. (C) U-VisualBERT~\cite{li2021unsupervised} extends oscar-style inputs into unpaired VLP and employs two separate branches to process text and image data. (D) VLMixer injects visual patches into the texts to form a ``multi-modal sentence", which is considered an intermediate representation to bridge the two modalities, since it keeps the syntactic structure of the original sentence meanwhile linking to the diverse visual patterns.}
    \vspace{-1em}
    \label{fig:relatedwork}

\end{figure}

Our key contributions are summarized as follows:
\begin{itemize}
    \item  We propose cross-modal CutMix to construct a multi-modal representation to bridge the images and texts, guiding the model to learn cross-modal alignment at the token level. 
    \item  We propose cross-modal contrastive learning upon CMC to facilitate instance-level alignments between unpaired images and texts, where semantically similar instances are pulled closer and dissimilar instances are pushed away.
    \item  {\color{black}Extensive experiments on diverse downstream tasks show that our approach achieves superior performance over previous unpaired VLP methods.}

\end{itemize}

\section{Related Work}

\paragraph{Paired vision-language pre-training.}
Benefiting from the soaring performance of transformers~\cite{vaswani2017attention} on representation learning in both computer vision and natural language processing~\cite{dosovitskiy2020image, devlin2019bert}, there is a surging interest in the field of joint pre-training~\cite{tan2019lxmert, li2019visualbert, li2020oscar} of parallel visual and language data. According to the learning objective, prior works can be divided into two categories, single-stream and dual-stream.
Single-stream models~\cite{tan2019lxmert, li2019visualbert, li2020oscar, chen2019uniter, kim2021vilt} aim to learn the joint representations of two modalities by a cross-modal encoder, which could handle very well the down-stream vision-language tasks with fine-level interactions and reasoning.
Dual-stream models~\cite{radford2021learning} learn separate representations for each modality by two independent uni-modal encoders, supervised by a constraint on the similarity between representations. It is suitable for downstream tasks requiring coarse-level cross-modal matching (e.g., image-text retrieval) and tasks where a single modality is presented, such as image and text classification.  

\paragraph{Unpaired vision-language pre-training.} Before the emergence of transformer-based VLP, image encoder and language encoder in traditional methods~\cite{yu2019deep} are pre-trained separately based on uni-modal datasets. However, there are no special designs for learning cross-modality alignments during pre-training, indicating that all knowledge about alignment is learned from the fine-tuning stage, where only a few manually-labeled image-text pairs are available. \citet{li2021unsupervised} explored the task of unpaired VLP, aiming to discover the complex interactions and semantic alignments between modalities in uni-modal pre-training datasets. Their method shares the encoders across modalities, forcing the samples in different modalities to be projected into the same space and thus encouraging alignments. Given an image, it concatenates the image regions together with their detector tags as aligned multi-modal inputs. Given a sentence, it directly considers the uni-modal subwords as input tokens. The pre-training objective is to reconstruct the masked inputs. We argue that, this scheme lacks the interactions between visual regions and linguistic cues (like quantifiers and words indicating relationships between two visual entities), resulting in a gap between pre-training and downstream tasks. Moreover, it lacks the ability to distinguish the alignment degree between visual and text data as no explicit matching supervision exists. Compared with~\citet{li2021unsupervised}, the most salient difference of VLMixer is that we use a cross-modal augmentation to construct semantic-invariant cross-modal inputs for 1) aligning the multi-modal and uni-modal view of the original sentence by contrastive learning; 2) effectively fusing the visual tokens and non-grounded linguistic tokens. Fig.~\ref{fig:relatedwork} summarizes mainstream paired and unpaired VLP methods\footnote{In this paper, unpaired VLP aims for cross-modal learning given image-only and text-only corpora. We classify some methods which rely on image-label pairs for pre-training a visual backbone into unpaired VLP since they use discrete concept categories instead of semantic-rich natural language with syntactic structure.}.

\begin{figure}[t]
    \centering
    \includegraphics[width=0.49\textwidth]{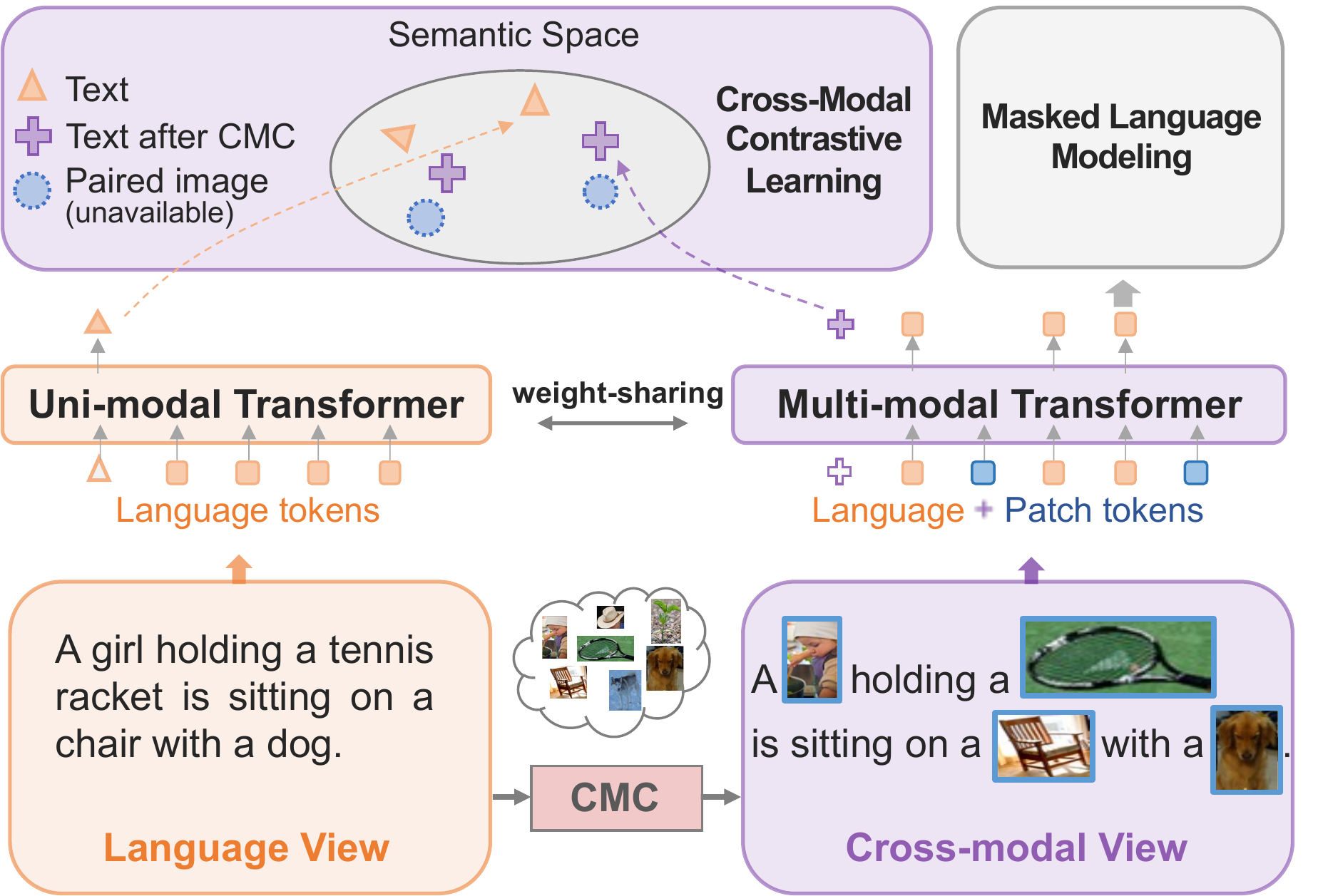}
    \caption{Visually-aided language pre-training. Given a sentence sample, we randomly wipe off some concept words in the sentence and then paste the visual patches with the same concept labels to obtain mixed sentences, serving as the cross-modal view of the original sentence. Two objectives are proposed for cross-modal learning: First, masked language modeling aims to learn the denoising representation, which encourages the token-level alignment between two modalities; Then, cross-modal contrastive learning guides the model to judge the instance-level alignment between the two views. Unlike contrastive learning used in paired VLP methods~\cite{li2020unimo,li2021align}, paired images are not available in our setting. The proposed contrast between text and text after CMC can be regarded as a proxy task of text-image contrast in paired VLP.}
    \label{fig:vall}
    \vspace{-1em}
\end{figure}

\paragraph{Unpaired image captioning.}
Unpaired image captioning focuses on training a useful image-to-text translation model without parallel image-text training data. Similar to unpaired VLP, the key component of this task is the cross-modal alignment. Existing literature designs different types of intermediate signals for connecting two modalities. \citet{gu2018unpaired} explored the pivot language to connect the source image and target language. \citet{feng2019unsupervised} explored an adversarial training framework including a concept detector and a sentence discriminator with three types of well-designed adversarial rewards, where concept words serve as the anchor points to bridge and align images and texts. \citet{gu2019unpaired} regarded the scene graph as an intermediate representation of each modality and trained a cycle-consistency adversarial method that maps scene graph features from the image to the text modality.

\paragraph{Data augmentation.}

{\color{black}{}
Data augmentation contains techniques for improving data diversity without collecting more data. It have been widely applied to several modalities, such as images~\cite{devries2017improved, yun2019cutmix}, texts~\cite{wei2019eda}, and audios~\cite{wei2019eda}. 
Our method is inspired by CutMix~\cite{yun2019cutmix} in vision tasks, which randomly removes image patches by overlaying salient patches from other images. The resultant image serves as an intermediate representation to bridge two images with different semantics. This paper constructs ``multi-modal sentences" to bridge the visual and linguistic modality, which could produce diverse multi-modal data without altering the semantics.
}

\section{VLMixer Pre-training}
VLMixer contains two parallel pre-training branches, visually-aided language pre-training (VALP) and tag-aided visual pre-training (TAVP). In VALP, given a  sentence sampled from the text-only dataset, we adopt cross-modal cutmix (CMC) to obtain a multi-modal view of the sentence and performs two learning objective on it, masked language modeling for reconstructing the masked inputs and contrastive learning for learning cross-modal alignments. 
In TAVP, given an image sampled from the image-only dataset, we follow~\cite{li2020oscar} to consider image tags and detected objects as the inputs for masked tag modeling. 
In the following, we introduce the cross-modal cutmix in subsection~\ref{sec:cmc}, the VALP and TAVP branches in subsections~\ref{sec:valp} and~\ref{sec:tavp}, respectively.

\subsection{Cross-Modal CutMix}
\label{sec:cmc}
The inputs in paired VLP~\cite{tan2019lxmert,li2020oscar} share a similar format with downstream tasks for fine-tuning: the mixed multi-modal sequence of both visual tokens and text tokens with consistent semantics. However, unpaired VLP without explicit alignments brings difficulties in constructing such a multi-modal input. Directly combining a text with a random image not only loses the cross-modal alignment but also introduces too much noise, which may overwhelm the interactions between intra-modal tokens. This section proposes cross-modal CutMix to construct diverse multi-modal sequences to mitigate the discrepancy between the pre-training and fine-tuning stages. 

\paragraph{Patch gallery.} We first collect a visual patch gallery of high-quality object regions with their concept labels from the image-only dataset. To this end, an off-the-shelf concept detector (e.g., Faster RCNN~\cite{ren2015faster}) is utilized to detect salient regions $x_i$ and predict their concept labels~${w^{\rm con}_i}$ and corresponding confidences~${c^{\rm con}_i}$. We denote the concept vocabulary as $\mathcal{C}$. Besides concepts of the current object, we also record ``contextual concepts", i.e., the concepts of other regions occurred in the same image, denoted as $\{(w^{\rm ctx}_{ij}, c^{\rm ctx}_{ij})\}$, where $w^{\rm ctx}_{ij}$ and $c^{\rm ctx}_{ij}$ represents the $j$-th contextual concept and its confidence score. The visual patches with their concepts are visually-grounded, serving as anchoring points to connect the images and sentences. We denote the patch gallery as: 
\begin{align}
\mathcal{G}=\left\{\left(w_i^{\rm con}, c^{\rm con}_{i}, \left\{\left(w_{ij}^{\rm ctx}, c^{\rm ctx}_{ij}\right)\right\}\right)\right\}.
\end{align}

\paragraph{CutMix visual patches into sentences.} Given a sentence $\mathbf{T} =\{w_n\}_{n=1}^{N}$ sampled from the text corpus $D^T$, our goal is to construct a multi-modal sequence while preserving the high-level semantics. For each (sub-)word in the sentence meanwhile appearing in the concept vocabulary $w_n \in \mathcal{C}$, we randomly replace it with a visual patch from the gallery with a probability of $r_{\rm cmc}$. The target visual patch is sampled from all patches with a concept label of $w_{n}$. We note that the sampled patches should not only accurately match the global semantics of the sentences, but also have diverse patterns for enhancing the generalization ability. This drives us to take the influence of the global semantics of the sentence into consideration. We design a context-aware sampling according to the following probability distribution. For a concept (sub-)word $w_n$ in $\mathbf{T}$, we calculate the probability of being chosen of all the items in the patch gallery. We sample a patch $x_q$ with ${q\sim {\rm Norm}(\{{p}_{i}\})}$ from the gallery, and $p_i$ is defined as:
\begin{align}
\begin{split}
{p}_{i} {\text{=}} \left \{
\begin{array}{ll}
    c^{\rm con}_i + \frac{r_{\rm ctx}}{|G_i|} \sum_{w^{\rm ctx}_{ij} \in {G_i}} c^{\rm ctx}_{ij}, &\text{\rm if } {w^{\rm con}_i=w_{n}} \\
    0,                            & {\rm otherwise}
\end{array}
\right.
\end{split},
\end{align}
where $G_i = \mathbf{T} \cap \{w^{\rm ctx}_{ij}\}$ represents the intersection between the sentence and contextual concepts of $x_i$, $\rm Norm(\cdot)$ normalizes the confidences $\{{p}_i\}$ as a probability distribution. $r_{ctx}$ controls the importance of contextual concepts for sampling. The resultant sequence $\mathbf{S}$ after CMC can be represented as a mixture of multi-modal tokens, like $\mathbf{S}=\{ w_1, x_{q_2}, w_3, x_{q_4},..., w_N\}$, where $x_{q_i}$ represents the sampled patch for the $i$-th (sub-)word.

\paragraph{K-shot CMC.}
Considering that a single patch only reveals a partial view of the concept (sub-)word, we propose the K-shot CMC which collects diverse patches as multiple views of this concept. Specifically, we replace $w_n$ with a set of patches that may come from different sources, by repeating the sampling process $K$ times. Thus, the resultant multi-modal tokens $\mathbf{S}$ becomes
$\{w_1, {x_{q_{2}^{(1)}},..., x_{q_{2}^{(K)}}}, w_3, {x_{q_{4}^{(1)}},..., x_{q_{4}^{(K)}}}, ..., w_N \}$.

\subsection{Visually-Aided Language Pre-training}\label{sec:valp}

VALP focuses on cross-modal learning from the text corpus with the assistance of the visual patch gallery. Different from U-VisualBERT~\cite{li2021unsupervised} which only adopts the uni-modal representation learning for text-only data, we construct the multi-modal inputs for effectively exploiting the multi-modal fusion by masked language modeling and cross-modal alignments by contrastive learning. The detailed illustration of VALP is shown in Fig.~\ref{fig:vall}. 

The input sentence $\mathbf{T}$ is firstly converted into a sequence of subwords $\{[CLS], w_1, w_2, ..., w_N, [SEP]\}$ by lower-case byte pair encoding (BPE)~\cite{sennrich2015neural}, where $[CLS]$ and $[SEP]$ denote the start and the end token of the subword sequence, respectively. We use cross-modal CutMix to obtain the cross-modal view $\mathbf{S}$. The representation of each patch token in $\mathbf{S}$ is the regional features produced by the concept detector. Then $\mathbf{S}$ is fed into a transformer encoder~\cite{vaswani2017attention} to learn cross-modal interactions by attention layers. 
The output feature vector of $[CLS]$ is regarded as the global representation of $\mathbf{S}$. 

\begin{figure}
    \includegraphics[width=0.45\textwidth]{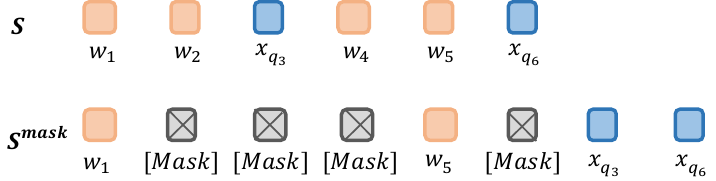}
    \caption{{\color{black} Masking strategy on the cross-modal view.}}
    \label{fig:cmc}
\end{figure}

\paragraph{Masked language modeling (MLM).}
We use a masking strategy analogy to BERT~\cite{devlin2019bert}. We randomly mask each language token in $\mathbf{S}$ with a probability of $15\%$. For each patch token, we add mask tokens into the sequence to indicate the position occurring CMC replacement. These mask tokens gather informative contextual features to recover the corrupted concept word at the same position. We denote the masked input as $\mathbf{S}^{\rm mask}$. We provide an example in Fig.~\ref{fig:cmc} to illustrate the difference between $\mathbf{S}$ and $\mathbf{S}^{\rm mask}$.

The goal of MLM is to reconstruct the original text from the two types of corruptions, i.e., cross-modal noise introduced by CMC and corruption from the masking mechanism. Thus, the model could effectively aggregate the contextual information and learn token-level alignments between visual and language tokens. The MLM objective is to minimize the negative log-likelihood of the reconstructed sequence $\hat{\mathbf{S}}$:
\begin{align}\label{loss_mlm}
\mathcal{L}_{\rm mlm}=-\mathbb{E}_{\mathbf{T} \sim \mathcal{D}^{\rm T}} \log(\hat{\mathbf{S}} = \mathbf{T}|\mathbf{S}^{\rm mask}).
\end{align}

\paragraph{Cross-modal contrastive learning (CMCL).} 
A common practice for paired VLP is the image-text matching task~\cite{chen2019uniter, li2020oscar}, where positive/negative samples, i.e., paired/unpaired inputs are constructed and the model is trained to distinguish whether the input image and text have similar semantics. Obviously, constructing such positive pairs requires well-aligned data and thus becomes unavailable for unpaired VLP. Despite the difficulties of finding a semantic-similar image for a given text, we propose to construct an intermediate representation by CMC that matches the meaning of the text. 

Given a training batch including a random set of the texts, we pair them with their CMC augmentation for contrastive learning, represented as
$\{(\mathbf{T}_{1},\mathbf{S}_{1}), \cdots, (\mathbf{T}_{M},\mathbf{S}_{M})\}$. 
For the anchor instance $\mathbf{T}_{m}$, we choose $\mathbf{S}_{m}$ as the positive instance and the remaining pairs in the batch as negative instances $\{\mathbf{S}_{l}\}_{l \neq m}$. The contrastive loss is calculated by:
\begin{align}\label{loss_cl}
    \mathcal{L}_{\rm cl} = - \sum_{m=1}^{M}
    \log \frac{\exp \left(f\left(\mathbf{T}^{\rm mask}_{m}, \mathbf{S}^{\rm mask}_{m}\right)/\tau \right)}
    { \sum_{l=1}^{M} \exp \left(f\left(\mathbf{T}^{\rm mask}_{m}, \mathbf{S}^{\rm mask}_{l}\right)/\tau \right)},
\end{align}
where $\mathbf{T}^{\rm mask}_m$ and $\mathbf{S}^{\rm mask}_l$ are the masked sequences of $\mathbf{T}_{m}$ and $\mathbf{S}_{l}$, $f(\mathbf{T}^{\rm mask}_m, \mathbf{S}^{\rm mask}_l)$ represents the cosine similarity between the output features on the $[CLS]$ tokens for $\mathbf{T}^{\rm mask}_m$ and $\mathbf{S}^{\rm mask}_l$. $\tau$ is the temperature ratio.

Note that our method differs from existing contrastive learning methods~\cite{radford2021learning, li2021align} for paired VLP for two reasons: 1) The paired image used in their model are unavailable in our setting; 2) {\color{black} The proposed contrast between uni-modal sample and multi-modal sample encourages multi-modal fusion, compared with the contrast between two uni-modal samples}.

\subsection{Tag-Aided Visual Pre-training}\label{sec:tavp}
TAVP mainly focuses on the exploitation of multi-modal knowledge from the visual-only data. Inspired by~\citet{li2021unsupervised}, we use image tags (concepts) as anchor points to connect vision and language, since they are detected from images but also play an important role in language learning. Specifically, given an image $\mathbf{I}$ from the image set $\mathcal{D}^{\rm I}$, a pre-trained concept detector is utilized to predict a number of image regions and their tags. The region token and the tag token are concatenated as the multi-modal representation of the image $\mathbf{Q}=\left(\mathbf{I}, \texttt{ Det}\left(\mathbf{I}\right)\right)$, where $\texttt{Det}(\mathbf{I})$ represents the sequence of tag tokens.

We adopt the mask-then-predict pre-training on image and tag tokens similar to OSCAR~\cite{li2020oscar}. Each tag token is randomly masked with a probability of $15\%$. Afterwards, we input the masked input $\mathbf{Q}^{\rm mask}$ into the transformer and calculate the reconstruction loss for the output sequence $\hat{\mathbf{Q}}$: 
\begin{align}\label{loss_mtm}
    \mathcal{L}_{\rm mtm} = -\mathbb{E}_{\mathbf{I} \sim \mathcal{D}^{\rm I}} \log(\hat{\mathbf{Q}} = \mathbf{Q}|\mathbf{Q}^{\rm mask}).
\end{align}

\begin{algorithm}[t]
\setstretch{1.0}
\caption{Unpaired VLP via CMC}
\label{alg:algo_pre}
\begin{algorithmic}
\STATE {\bfseries Input:} image set $D^I$, text set $D^T$.
\STATE {\bfseries Output:} pre-trained transformer.
{\color{black}
\STATE Construct a patch gallery $\mathcal{G}$ from $D^I$}

\FOR{$iter$ := 1 {\bfseries to} $max\_iter$}
    \STATE Sample a mini-batch of sentences $\mathbf{T}$ from $D^T$.
    \STATE Sample a mini-batch of images $\mathbf{I}$ from $D^I$.
    \STATE Obtain $\mathbf{S}$ by performing CMC on $\mathbf{T}$.
    \STATE Obtain image tokens with tags $\mathbf{Q} = \left(\mathbf{I}, \texttt{ Det}\left(\mathbf{I}\right)\right)$.
    \STATE Obtain $\mathbf{T}^{\rm mask}$, $\mathbf{S}^{\rm mask}$, $\mathbf{Q}^{\rm mask}$ by random masking.
    \STATE $\mathbf{\hat{T}} = \texttt{Transformer}\left(\mathbf{T}^{\rm mask}\right)$
    \STATE $\mathbf{\hat{S}} = \texttt{Transformer}\left(\mathbf{S}^{\rm mask}\right)$
    \STATE $\mathbf{\hat{Q}} = \texttt{Transformer}\left(\mathbf{Q}^{\rm mask}\right)$ 
    \STATE Compute $\mathcal{L}_{\rm total}$ with \eqref{loss_total} and update model parameters.
    \ENDFOR
\end{algorithmic}
\label{algo:uvlp}
\end{algorithm}

\subsection{Training Objective}
The overall training objective is defined as follows:
\begin{align}\label{loss_total}
    \mathcal{L}_{\rm total} = \mathcal{L}_{\rm mlm} + \mathcal{L}_{\rm cl} + \mathcal{L}_{\rm mtm},
\end{align}
which is the summation of the masked language modeling loss, contrastive loss, and the masked tag modeling loss. At each iteration, we sample a mini-batch of images and a mini-batch of texts for loss calculation. The detailed pre-training process is summarized in Algorithm~\ref{algo:uvlp}. 

\section{Experiments}

For fair comparisons, we first follow the standard practice in unpaired vision-language (VL) tasks~\cite{li2021unsupervised, feng2019unsupervised} to evaluate the model performance on the paired VL datasets without the alignment information. Next, we show that VLMixer could benefit from large-scale images or texts collected independently from different sources. Finally, we conduct ablation studies on important design choices to show the effectiveness of VLMixer.

\subsection{Datasets}\label{sec:dataset}

We use a variety of datasets covering diverse visual and language patterns. Specifically, three kinds of pre-training datasets are taken into account: image-text pairs, image-only collections and text-only corpora. The {paired VL datasets} contain COCO Captions~\cite{lin2014microsoft}, Visual Genome~\cite{krishna2017visual}, Conceptual Captions 3M~\cite{sharma2018conceptual}, SBU Captions~\cite{ordonez2011im2text}, Flickr30K~\cite{plummer2015flickr30k}, and GQA~\cite{hudson2019gqa}, totally 4.1M images and 5.6M captions. For {additional image-only data}, we use Open Images\cite{kuznetsova2018open} containing 1.7M images. The {text-only corpus} comes from three sources: 1) human-annotated captions from existing video captioning datasets, i.e., MSVD~\cite{chen2011collecting}, MSRVTT~\cite{xu2016msr}, VATEX~\cite{wang2019vatex}, and ActivityNet Captions~\cite{krishna2017dense}; 2) Auto-crawled captions from a online stock photography website Shutterstock, provided by~\citet{feng2019unsupervised}; 3) General text segments from BookCorpus~\cite{zhu2015aligning}. The total size of text-only instances is around 15.5M. Detailed statistics of the pre-training corpus are provided in Table~\ref{tab:dataset}.

\begin{table}[t]
\small
    \centering
    \begin{spacing}{1.2}
    \setlength{\tabcolsep}{ 1.2 mm}{
    \begin{tabular}{c | c | c | c}
    \toprule
         Dataset & Images & Texts & Text Domain \\
     \midrule
         COCO (train) & 112K & 560K & Image Caption\\
         Conceptual Captions (train) & 3M & 3M & Image Caption \\
         SBU Caption (all) & 840K & 840K & Image Caption\\
         Flickr30k (train) & 29K & 145K & Image Caption\\
         VQA (train) & 83K & 445K & Question \\
         GQA (train) & 79K & 1.0M & Question \\
         VG-QA (train) & 87K & 931K & Question \\
         MSVD (train) & - & 48K & Video Caption \\
         MSRVTT (train) & - & 130K & Video Caption \\
         VATEX (train) & - & 260K & Video Caption\\
         ActivityNet Captions (train) & - & 36K & Video Caption \\
       Shutterstock (all) & - & 1M & Caption\\
         BookCorpus & - & 14M & General Text\\
         OpenImages (od train) & 1.67M & - & - \\
     \bottomrule
    \end{tabular}
    }
    \end{spacing}
    \caption{Pre-training dataset statistics. We use several large-scale image and text datasets with diverse language patterns.}
    \label{tab:dataset}
\end{table}

\begin{table*}[]
    \small
    \centering
    \setlength{\tabcolsep}{ 0.76 mm}{
    \begin{spacing}{1.2}
    \begin{tabular}{lcc  c  cc  ccc ccc  c  }
    \toprule
         \multirow{2}{*}{Method} & \multicolumn{2}{c}{Pre-training Data} & \multicolumn{1}{c}{VQA} & \multicolumn{2}{c}{NLVR$^2$} & \multicolumn{3}{c}{Text Retrieval} & \multicolumn{3}{c}{Image Retrieval} & GQA  \\
         & Image&Text & Test-Dev & Dev & Test & \ R@1 & R@5 & R@10 & R@1 & R@5 & R@10 & Test-Dev  \\
         \midrule 
        \textbf{Paired VLP} \\
            \ \ UnicoderVL$_{\rm base}$~\cite{li2019unicoder}  &\multicolumn{2}{c}{  }& - & - & - & 62.3 & 87.1 & 92.8 & 46.7 & 76.0 & 85.3 & -\\
            \ \ UNITER$_{\rm base}$~\cite{chen2019uniter} && & 72.27 &  77.14  & 77.87 & 63.3 & 87.0 & 93.1 & 48.4 & 76.7 & 85.9 & -  \\
            \ \ OSCAR$_{\rm base}$~\cite{li2020oscar}  && &  73.16 & 78.07 & 78.36 & 70.0 & 91.1& 95.5& 54.0& 80.8& 88.5 &61.58   \\
            \ \ VILT$_{\rm base}$~\cite{kim2021vilt} && & 71.26 &  75.70  & 76.13 & 61.5 & 86.3 & 92.7 & 42.7 & 72.9 & 83.1 & -\\
            \ \ VinVL$_{\rm base}$~\cite{zhang2021vinvl}  && & 75.95 & 82.05 & 83.08 & 74.6 & 92.6 & 96.3 & 58.1 & 83.2 & 90.1 & 65.05   \\
            \ \ ALBEF~\cite{li2021align} & & & 75.84 & 82.55 & 83.14 & 77.6 & 94.3 & 97.2 & 60.7 & 84.3 & 90.5 & -\\
            \midrule \midrule
            \textbf{Unpaired VLP} \\
            \ \ BERT$_{\rm base}$~\cite{devlin2019bert} & None & None & 64.85 &  51.30 & 51.34 & 57.44 & 84.00 & 91.58 & 44.03 & 74.12 & 84.06 & 50.20 \\
            \ \ VinVL$_{\rm unpaired}$~\cite{zhang2021vinvl} & COCO & COCO &  71.78 & 71.14 & 72.01  & 61.92 & 86.90 & 93.08 & 46.90 & 76.18 & 85.53 & 62.24 \\
            \ \ U-VisualBERT~\cite{li2021unsupervised}*   & COCO & COCO &  72.41 & - & - & - & - & - & - & - & - & -  \\
            \rowcolor{Gray}
            \ \ {VLMixer}  & COCO & COCO & \textbf{72.60} & \textbf{72.71} & \textbf{73.08} & \textbf{62.69}  & \textbf{87.35} & \textbf{93.64} & \textbf{47.95} & \textbf{77.06} & \textbf{86.22} & \textbf{63.13} \\
                
            \midrule
            \ \ U-VisualBERT~\cite{li2021unsupervised}   & CC3M & CC3M+BC &  70.74 & 71.74 & 71.02 & - & - & - & - & - & - & -  \\
            \ \ VinVL$_{\rm unpaired}$~\cite{zhang2021vinvl}  & CC3M & CC3M &  72.20 & 68.96 & 68.94 & 62.08 & 86.04 & 93.00 & 47.29 & 76.15 & {85.53}& 63.12 \\
            \rowcolor{Gray}
            \ \ {VLMixer} & CC3M & CC3M & {72.66} & {74.31} & {73.86} & {62.20} &{86.32} & 92.80 & {47.44} & {76.22} & 85.41 & 62.65 \\ 
            \rowcolor{Gray}
            \ \ \textcolor{black}{{VLMixer}} & Full  & Full  & \textbf{72.89}  & \textbf{76.61}  & \textbf{77.01} & \textbf{64.76} & \textbf{88.56} &\textbf{ 94.22} & \textbf{50.06} & \textbf{78.36} & \textbf{86.91} & \textbf{63.25}  \\
     \bottomrule
    \end{tabular}
    \end{spacing}
    }
    \caption{Comparison with state-of-the-art unpaired VLP methods. We report the performance on COCO and CC3M for a fair comparison with previous state-of-the-art methods. ``Full" data means we leverage all image data and text data introduced in subsection~\ref{sec:dataset}. ``CC3M" and ``BC" denote the conceptual captions 3M and the BookCorpus datasets. $^*$ denotes the results of our re-trained model with the VinVL object features. We also list the performance of paired VLP methods for reference.} 
    \label{tab:UnpairSota}
\end{table*}

\begin{table*}[t]
\small
    \centering
    \setlength{\tabcolsep}{0.8 mm}{
    \begin{spacing}{1.2}
    \begin{tabular}{ccc  c  c  cc  ccc  ccc  }
    \toprule
         \multicolumn{3}{c}{VALP} & \multirow{2}{*}{TAVP} &\multicolumn{1}{c}{VQA} & \multicolumn{2}{c}{NLVR$^2$}& \multicolumn{3}{c}{Text Retrieval} & \multicolumn{3}{c}{Image Retrieval}\\
         MLM & CMC & CMCL & & Test-Dev & Dev & Test & R@1  & R@5 & R@10  & R@1  & R@5  & R@10\\
         \midrule
         &&&$\surd$ & 71.16 & 70.52 & 69.23 & 60.18 & 85.50 & 91.72 & 45.87 & 75.39 & 84.96  \\
        $\surd$&&&  & 71.50 & 50.89 & 52.16  & 49.32 & 78.02 & 87.72 & 38.04 & 69.62 & 80.92  \\
        $\surd$&&& $\surd$ & 72.00 & 72.52 & 72.20 & 59.30 & 85.36 & 91.76 & 45.78 & 74.94 & 84.60 \\
        $\surd$&$\surd$&&  & 71.52 & 71.13 & 70.99 & 60.40 & 85.72 & 92.92 & 46.92 & 75.86 & 85.31 \\
        $\surd$&$\surd$&& $\surd$ & 71.84 & \textbf{73.19} & 72.81 & 60.54 & 86.24 & 92.44 & {47.29} & 76.43 & {85.61} \\
    \rowcolor{Gray}
        $\surd$ & $\surd$ & $\surd$ & $\surd$ & $\textbf{72.60}_{\pm \textbf{0.10}}$ & ${72.71}_{\pm 0.61}$ & $\textbf{73.08}_{\pm \textbf{0.26}}$ & $\textbf{62.69}_{\pm \textbf{0.51}}$ & $\textbf{87.35}_{\pm \textbf{0.19}}$ & $\textbf{93.64}_{\pm \textbf{0.14}}$ & $\textbf{47.95}_{\pm \textbf{0.21}}$ & $\textbf{77.06}_{\pm \textbf{0.13}}$ & $\textbf{86.22}_{\pm \textbf{0.08}}$ \\
        \midrule
        \multicolumn{4}{c}{Paired Pre-training} &  72.39& 75.28& 75.54& 65.10& 88.82& 94.38& 50.23& 78.49& 87.13 \\
     \bottomrule
    \end{tabular}
    \end{spacing}
    }
\caption{Ablation studies of pre-training objectives. All models are pre-trained on COCO without alignment information except in the last row. For paired pre-training, we feed the concatenation of image, tag, and language tokens into the transformer and use image-text matching loss with masked token modeling loss as the training objectives, as in~\cite{li2020oscar}. For the final model, we run three times to report the mean and standard deviation.}
    \label{tab:ablation}
\end{table*}

\subsection{Experimental Setting}

\paragraph{Implementation details.} We use a Base Transformer with 12 layers of transformer block and a hidden size of 768 as the backbone. For position embedding, we adopt the learnable position embedding for language/tag tokens and use a linear projection of spatial positions for patch/image tokens. To reduce the computation cost, we restrict the max token length in TAVP and VALP to 100 and 80, respectively. For VALP, we first collect a subset of object regions within the image dataset as the patch gallery by filtering regions with high confidence. The object regions are detected by an off-the-shelf concept detector ResNeXt-152 C4 provided by~\citet{zhang2021vinvl}. The size of the concept vocabulary is 1600. For each sentence, we adopt K-shot CMC to enhance the diversity of sampled patch tokens with $K$=15. The replacing probability $r_{cmc}$ in CMC is set to 0.5 and the context weight $r_{ctx}$ is set to 0.5.
Note that we followed the standard practices in unpaired VL tasks to prevent selecting patches in the paired image.
To reduce the noisy data, we drop the sentences shorter than five words as there is a high probability that it does not contain concept words. The temperature ratio $\tau$ in CMCL is set to 0.1.

\paragraph{Pre-training and fine-tuning.}
We initialize VLMixer from the parameters of BERT$_{\rm base}$, and pre-train the model on unpaired image and text data for a maximum of 300k steps. An Adam optimizer is adopted with an initial learning rate of 5e-5 and a mini-batch size of 1024. The warm-up rate is set to 10\%. Mixed precision training is used to accelerate the training stage. The training time on the full pre-training data is around six days on 16 Telsa A100 GPUs. 

After pre-training, we adapt the weights of VLMixer to five downstream tasks, i.e., VQA~\cite{goyal2017making}, NLVR$^{2}$~\cite{suhr2018corpus}, image retrieval (COCO 5K), text retrieval (COCO 5K), and GQA~\cite{hudson2019gqa}. We follow the fine-tuning strategy and  evaluation metrics in~\citet{zhang2021vinvl}  for downstream tasks.

\subsection{Comparison with State-of-the-Art Methods}

We compared VLMixer with the following methods in the setting of unpaired pre-training: 1) \textbf{U-VisualBERT} is a pioneer work in unpaired VLP. It uses a parallel pre-training scheme for each modality and utilizes tags as anchor points to connect images and texts. 2) \textbf{VinVL$_{\rm unpaired}$}: We modified the paired VLP method VinVL to fit the unpaired setting by simply considering a text with randomly sampled images as inputs for mask-then-predict learning. The image-text matching loss is disabled. 3) \textbf{BERT$_{\rm base}$} is the standard BERT base model pre-trained on text datasets.

The performance comparison is shown in Table~\ref{tab:UnpairSota}. We test the performance of VLMixer based on pre-training corpora with three scales, COCO, CC3M, and the full corpus. Our method achieves a better performance on most downstream tasks than other methods under a similar size of pre-training data. Compared with paired VLP, we achieve comparable performance with UNITER$_{\rm base}$ and VILT$_{\rm base}$, showing that pre-training on large-scale easy-to-collect unpaired data has great potential to benefit the vision-language tasks.

As images and captions in COCO/CC3M datasets come from the same source, it is natural to ask what if image and text sets are not fully aligned. Then we conduct pre-training on full corpora, which contains rich images and diverse language patterns that are collected from different sources. The language data contains image caption, video caption, question, and general text, while the image data are usually for common image recognition.
The superior performance of VLMixer on the full pre-training data shows that our model could effectively learn useful cross-modal interactions from large-scale images or texts independently collected from different sources.

\subsection{Ablation Studies}
\paragraph{Main results.}
The ablation study of the proposed method is shown in Table~\ref{tab:ablation}. ``MLM+CMC" achieves a considerable performance-boosting over ``MLM", which means the introduction of patch tokens significantly boosts the learning of cross-modal interactions. We also notice that ``MLM+CMC" performs better than TAVP, which shows the importance of the syntactic information introduced by multi-modal sentences.
When incorporating tag-aided visual pre-training (TAVP), the overall performance could obtain further improvement. 
It is an interesting phenomenon that our best model, which uses CMC and CMCL together, could achieve better performance than the paired pre-training model on VQA. We conjecture that the reason may lie in that mixing the language tokens with patch tokens could largely increase the data diversity, which could benefit the model to achieve good generalization performance.

\begin{figure}
    \centering
    \vspace{-1em}
    \includegraphics[width=0.48\textwidth]{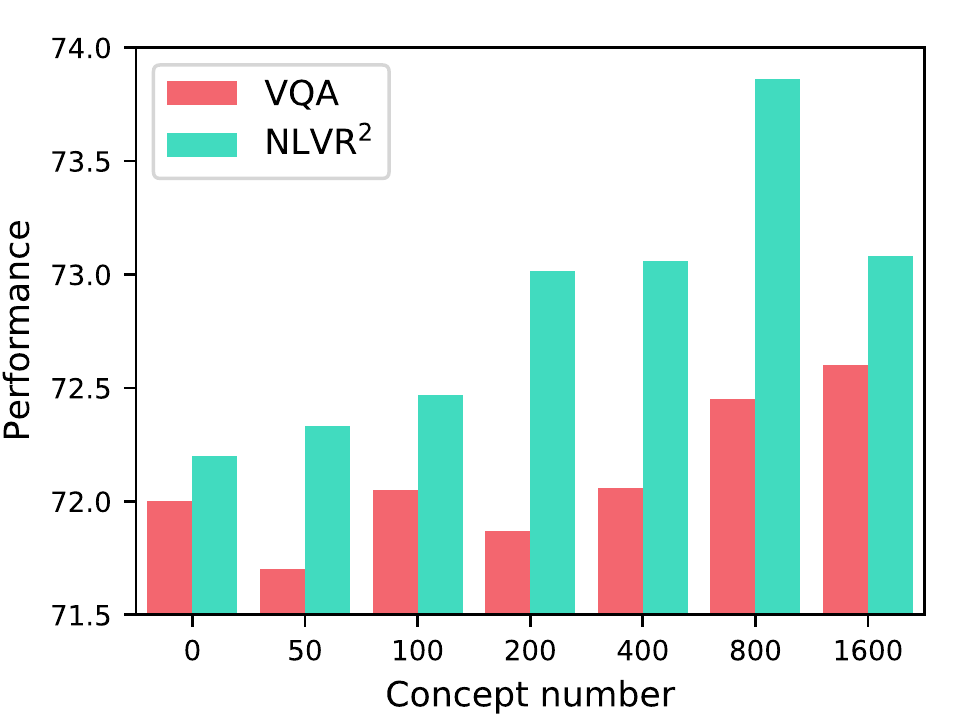}
    \caption{The downstream performance using different number of concepts in the patch gallery. We modulate the number of shared concepts by controlling the concept number of the patch gallery.}
    \label{fig:ablationCMC}
\end{figure}

\paragraph{Cross-modal CutMix.} The number of shared concepts between patch gallery and text corpus matters, since it reflects the global alignment between an image dataset and a text dataset. In Fig.~\ref{fig:ablationCMC}, we test the influence of the number of shared concepts by constructing six subsets of COCO Captions with an increasing number of concepts. We see that with the increase of concept number, the NLVR$^2$ performance gradually improves. The reason for the slight decrease at 1600 concepts may lie in too many concepts introducing inevitable more misrecognition, especially for long-tailed concept classes. {\color{black} We conclude that the model could benefit from an increasing number of shared concepts, showing that VLMixer effectively exploits the multi-modal interactions and makes better usage of the concepts.}

We also verify the effectiveness of two techniques, K-shot CMC and context-aware sampling. For K-shot CMC, a larger $K$ performs better. A small $K$ causes an imbalance between language and patch tokens, resulting in pre-training being dominated by language tokens, which in turn restricts cross-modal learning. For context-aware sampling, combining confidences from concept tokens and context tokens performs better than removing the context tokens, indicating that context tokens are informative for accurate sampling.

\begin{table}[h]
\small
\renewcommand\arraystretch{1.0}
\setlength{\tabcolsep}{1mm}{
\begin{spacing}{1.2}
\begin{tabular}{lccc}
    \toprule
 \multirow{2}{*}{Method}  & \multicolumn{1}{c}{VQA} & \multicolumn{2}{c}{NLVR$^2$}\\
         & Test-Dev & Dev & Test \\
    \midrule
    w/o K-shot CMC ($K$=1) & 71.88 & 71.77 & 72.25 \\
    w/o context-aware sampling ($r_{\rm ctx}$=0) & 71.86 & 72.07 & 72.48 \\
    \rowcolor{Gray}
    {VLMixer} &\textbf{ 72.60} & \textbf{72.71} &  \textbf{73.08} \\
    \bottomrule
\end{tabular}
\end{spacing}
}
\caption{Ablation of Cross-modal CutMix.}
\label{table:abl}
\end{table}

\paragraph{Contrastive learning.} To effectively learn the cross-modal alignment, we propose the contrast between uni-modal textual sentences and the multi-modal sentences after CMC. We compare this design with the previous data augmentation method used in contrastive learning. Specifically, we implement two text data augmentation methods: 1) Crop $k\%$. We randomly crop the sentence and keep a continuous segment with a length of $100$-$k\%$. 2) Delete $k\%$. We randomly remove $k\%$ words from the sentence. The performance of the ablated methods is shown in Table~\ref{tab:ablationCL}. Compared with the model without contrastive learning, our method could improve both VQA and NLVR$^2$, while text-text contrast can not achieve consistent improvement on two tasks. Our method is superior to text-text contrast. The reason may be that our method encourages the cross-modal fusion of inputs where two modalities are semantically consistent, and discourages that where two modalities are semantically incompatible.

\begin{table}[t]
\small
    \centering
    \setlength{\tabcolsep}{2 mm}{
    \begin{spacing}{1.2}
    \begin{tabular}{l  c  cc   }
    \toprule
         \multirow{2}{*}{Method}  & \multicolumn{1}{c}{VQA} & \multicolumn{2}{c}{NLVR$^2$}\\
         & Test-Dev & Dev & Test \\
         \midrule
       \rowcolor{Gray}
    \textbf{Cross-modal contrast} & {72.60} & 72.71 & {73.08} \\
        w/o contrastive learning & 72.00 & 72.52 & 72.20 \\
        Text-text contrast: crop $10\%$ & 72.04 &71.82 & 72.74\\
        Text-text contrast: crop $20\%$ & 71.79 & {72.97} & 72.15\\
        Text-text contrast: crop $30\%$  & 72.52& 70.15 & 70.17 \\
        Text-text contrast: delete $10\%$  &72.70 &71.54 & 71.42 \\
        Text-text contrast: delete $20\%$ & 71.71&71.60 & 71.51 \\
        Text-text contrast: delete $30\%$ & 71.77 &72.31 & 72.27 \\
     \bottomrule
    \end{tabular}
    \end{spacing}
    }
    \caption{Ablation study of the contrastive learning methods and data augmentations. All models are pre-trained on COCO.}
    \label{tab:ablationCL}
\end{table}

\section{Conclusions}
This paper presents a new method named VLMixer for unpaired vision language pre-training. Different from traditional methods that use tags as the anchor to bridge the two modalities, we propose to construct the cross-modal view of the textual sentences by cross-modal CutMix. By doing so, the diversity of multi-modal data could be increased to a large extent without altering the semantics. Furthermore, to enable better alignment learning at the instance level, we build the contrastive learning objective on multi-modal sentences to pull together semantically similar instances and push away semantically dissimilar instances. Experiments on five downstream tasks show that our method achieves state-of-the-art performance in unpaired VLP. The ablation studies of the design choices in CMC and contrastive learning verify the effectiveness of the proposed model.

\section{Acknowledgement}
This work is supported by the National Natural Science Foundation of China under Grant No.~61972188, No.~62122035, No.~61906081, No.~62106097, and the China Postdoctoral Science Foundation (2021M691424). Ping Luo is supported by the General Research Fund of HK No.~27208720 and 17212120. 

\nocite{langley00}

\bibliography{main}
\bibliographystyle{icml2022}



\end{document}